\documentclass[journal,twoside,web]{ieeecolor}
\usepackage{tmi}
\usepackage{cite}
\usepackage{amsmath,amssymb,amsfonts}
\usepackage{graphicx}
\usepackage{textcomp}

\usepackage[ruled]{algorithm2e}
\usepackage{algpseudocode}

\usepackage{subfig}

\usepackage{url}

\begin{document}

\bstctlcite{IEEEexample:BSTcontrol}

\title{Active learning using adaptable task-based prioritisation}
\author{
Shaheer U. Saeed, 
João Ramalhinho, 
Mark Pinnock, 
Ziyi Shen, 
Yunguan Fu, 
Nina Montaña-Brown, 
\\Ester Bonmati, 
Dean C. Barratt, 
Stephen P. Pereira, 
Brian Davidson, 
Matthew J. Clarkson, 
Yipeng Hu 
\thanks{S. U. Saeed, J. Ramalhinho, M. Pinnock, Z. Shen, Y. Fu, N. Montaña-Brown, E. Bonmati, D. C. Barratt, M. J. Clarkson and Y. Hu are with the Centre for Medical Image Computing, Wellcome/EPSRC Centre for Interventional and Surgical Sciences, Department of Medical Physics and Biomedical Engineering, University College London, London WC1E 6BT, U.K. (e-mail: shaheer.saeed.17@ucl.ac.uk, joao.ramalhinho.15@ucl.ac.uk, mark.pinnock.18@ucl.ac.uk, ziyi-shen@ucl.ac.uk, yunguan.fu.18@ucl.ac.uk, nina.brown.15@ucl.ac.uk, e.bonmaticoll@westminster.ac.uk, d.barratt@ucl.ac.uk, m.clarkson@ucl.ac.uk, yipeng.hu@ucl.ac.uk) }
\thanks{Y. Fu is also with InstaDeep, London W2 1AY, U.K. (e-mail: yunguan.fu.18@ucl.ac.uk) }
\thanks{E. Bonmati is also with the School of Computer Science and Engineering, University of Westminster, London W1W 6UW, U.K. (e-mail: e.bonmaticoll@westminster.ac.uk) }
\thanks{S. P. Pereira is with the Institute for Liver and Digestive Health, University College London, London NW3 2QG, U.K. (e-mail: stephen.pereira@ucl.ac.uk) }
\thanks{B. Davidson is with the Division of Surgery and Interventional Sciences, University College London, London WC1E 6BT, U.K. (e-mail: b.davidson@ucl.ac.uk) }
}

\maketitle

\begin{abstract}
Supervised machine learning-based medical image computing applications necessitate expert label curation, while unlabelled image data might be relatively abundant. Active learning methods aim to prioritise a subset of available image data for expert annotation, for label-efficient model training. We develop a controller neural network that measures priority of images in a sequence of batches, as in batch-mode active learning, for multi-class segmentation tasks. The controller is optimised by rewarding positive task-specific performance gain, within a Markov decision process (MDP) environment that also optimises the task predictor. In this work, the task predictor is a segmentation network. A meta-reinforcement learning algorithm is proposed with multiple MDPs, such that the pre-trained controller can be adapted to a new MDP that contains data from different institutes and/or requires segmentation of different organs or structures within the abdomen. We present experimental results using multiple CT datasets from more than one thousand patients, with segmentation tasks of nine different abdominal organs, to demonstrate the efficacy of the learnt prioritisation controller function and its cross-institute and cross-organ adaptability. 
We show that the proposed adaptable prioritisation metric yields converging segmentation accuracy for the novel class of kidney, unseen in training, using between approximately 40\% to 60\% of labels otherwise required with other heuristic or random prioritisation metrics. For clinical datasets of limited size, the proposed adaptable prioritisation offers a performance improvement of 22.6\% and 10.2\% in Dice score, for tasks of kidney and liver vessel segmentation, respectively, compared to random prioritisation and alternative active sampling strategies. 
\end{abstract}


\section{Introduction}
\label{sec:introduction}

Medical imaging tasks are increasingly being automated using machine learning by utilising expert annotated data \cite{lee_deep_med, erickson_ml_med}. Supervised learning using expert annotations allows for reliable predictions from the trained model, however, this expert annotation may often be expensive. Applications such as complex surgical planning thus become challenging to develop, due to the need for many structures to be annotated at the voxel-level and different regions of interest (ROIs) required by subsequent procedures mandated by local expertise and protocols. This is further complicated by the now well-known problem of generalisation from deep models across different institutions, all of which are often under data size constraints.

Active learning (AL) aims to directly address the expensive data labelling by prioritising a subset of available unlabelled data for annotation, such that the machine learning models trained with these annotated data reach a predefined, or the same, performance level with fewer labelled samples, as models trained with all data being labelled. The efficiency of the performance convergence measures performance of the AL methods, in terms of the quantity of annotated data, i.e. required number of \textit{AL iterations}, often compared with random sampling without prioritisation \cite{budd_active_survey, settles_active}. Therefore, metrics that valuate how data samples affect AL convergence (hereinafter referred to as prioritisation metrics) are the key to the goal of fast convergence, i.e. using as few labelled samples as possible. Informativeness and representativeness are regarded as the main criteria in existing prioritisation metrics \cite{budd_active_survey}. 

Informativeness estimates information gained if a particular labelled sample is added to training. Uncertainty with respect to the given samples is often used to quantify the informativeness, as it measures the amount of uncertain, therefore likely unknown, information that could be learnt by including the samples. For tasks like image segmentation, a summation of the lowest class probabilities over all pixels can be used \cite{shannon_active, lewis_active_text}, while high class probabilities are assumed high prediction confidence.
An ensemble of multiple models was proposed for quantifying uncertainty \cite{settles_active, czolbe_segmentation_uncert}. 
Monte-Carlo Dropout-based uncertainty estimation \cite{gal_mcdropout} was also proposed and may be viewed as a special case of ensemble methods. 

Representativeness measures the similarity between data samples, such that an effective AL strategy can be designed for prioritising those samples that can efficiently represent many others \cite{budd_active_survey}. Distances between multiple images have been proposed, for example, based on features extracted from a trained model for a different, usually unsupervised, task such as self-reconstruction \cite{yang_rep_quant, Smailagic_rep_quant, ozdemir_rep_quant}. Representativeness can also be combined with informativeness measures \cite{yang_rep_quant, Smailagic_rep_quant, ozdemir_rep_quant}.

%The underlying assumption for AL metrics, such as those described above, is that annotations are unambiguous and noise-free \cite{czolbe_segmentation_uncert}. The impact of an annotated sample on the learning system is not quantified post-annotation due to the fixed and non-adaptive nature of the prioritisation metrics. 
%Therefore, AL metrics, such as those described above, do not account for ambiguity and noise within annotations since feedback from the learning system is not incorporated post-annotation. 

However, general prioritisation metrics, such as Monte-Carlo Dropout and ensemble, have shown nearly equivalent performance to random sampling \cite{czolbe_segmentation_uncert}. The fixed and non-adaptive nature of these metrics could lead to adverse consequences. For example, high uncertainty in samples may in fact be a result of label error or inconsistency, due to manual annotation difficulty \cite{czolbe_segmentation_uncert}. It has been speculated that not accounting for the impact of annotated samples, \textit{post annotation}, and assuming that annotations are unambiguous and noise-free have led to the ineffective prioritisation metrics \cite{czolbe_segmentation_uncert, meng_rl_active}. This has been consistent with our preliminary results in a task of segmenting kidney on 3D CT images (summarised in Fig. \ref{fig:ablat}, with further details discussed in Sec. \ref{sec:exp}). In contrast, task-based prioritisation can utilise task-specific feedback in formulating the prioritisation, such as the performance of a trained model for the subsequent task. This task-based feedback enables post annotation impact to be measured during model training \cite{meng_rl_active, woodard_one_shot} and may alleviate the discussed limitations for individual tasks.

%\subsection*{AL for multi-organ and multi-institute medical images} % difficult problem with imperfect annotations

% Medical images and their annotation is not only expensive but may have high variability between observers, even for experienced observers \cite{sudre_multi_rater, czolbe_segmentation_uncert, chalcroft_multi_obs}. In such cases, where annotations may be ambiguous, assuming noise-free and unambiguous annotations and not accounting for the impact of labelled samples on the learning system, for AL prioritisation metric design could be detrimental \cite{czolbe_segmentation_uncert}. Organ boundary delineation on medical images, such as computed tomography (CT) and magnetic resonance (MR) images, may have high variability and ambiguity in annotations \cite{czolbe_segmentation_uncert, Montagne_mr_multi_obs}. 

% As demonstrated by Czolbe et al. \cite{czolbe_segmentation_uncert}, for medical imaging tasks such as organ boundary delineation, AL metrics that do not incorporate task-based feedback, such as Monte-Carlo Dropout-based uncertainty estimation or ensemble methods, are unable to outperform random sampling strategies. Based on preliminary experiments for a task of kidney segmentation on 3D CT images (see Sec. \ref{sec:exp} for further details on the dataset used), we had similar findings. The performance of the AL system that uses Monte-Carlo Droupout-based uncertainty estimation, compared with a random sampling system, is presented as a plot in Fig. \ref{fig:ablat} (see Sec. \ref{sec:exp} for further details on the AL experiment setup). 

In this work, we focus on organ segmentation on 3D abdominal CT images. Multiorgan segmentation has a number of clinical applications \cite{fu_multi_organ_review, gibson_multi_organ_seg}. Planning laparoscopic liver resection or liver surgery in general is one such example, in which localising the liver, liver vessels and surrounding anatomy is necessary for existing inter-modality image registration \cite{ramalhinho_liver} and useful for subsequent navigation \cite{fusaglia_liver} during the procedure. Moreover, AL will greatly benefit the development of automatic segmentation models for different clinical requirements, because of the potentially diverging protocol-specific needs, such as the types of vessels and/or organs required for different registration algorithms and changing local image-navigating procedures. We thus identify two aspects for a desirable AL approach in this application: 1) prioritising CT images to be annotated for the required ROI types (organs or anatomical structures), potentially new and unseen in developing such prioritisation strategy, and 2) the ability to adapt or generalise such prioritisation to image data from a different and novel institute. 
%These may be considered of practical value in clinical adoption and deployment of deep learning-based segmentation tools.   
%in particular, the AL potentially can help alleviate the lack of training data for pre-operative planning applications where the appropriate cohort is limited by the number of available surgical procedures, such as those who have undergone laparoscopic liver resection. We aim to develop an AL system for the task of organ segmentation on these images. 

We first propose a prioritisation metric based on direct feedback from the segmentation task using annotated samples, which is learnt using reinforcement learning (RL) based meta-learning. Second, we outline a mechanism, using the proposed meta-RL, to allow for the metric to be adapted to new data distributions including data from new institutes and for segmenting new ROI classes i.e. organs or structures unseen in training. In our formulation, task-based feedback for AL is delivered by means of a reward signal in the RL algorithm, in order to learn a prioritisation metric function. The reward signal is computed by measuring performance of a partially trained model on a set of samples for which annotations are available. The meta-RL further enables such prioritisation function to be useful across wider domains than with ``simple'' RL \cite{duan_metarl, wang_metarl, botvinick_metarl}. 

It is important to highlight the difference between the proposed prioritisation metrics for AL and few-shot learning, which requires small number of annotated data from the novel classes and/or institutions during adaptation, e.g. \cite{li_fewshot, feyjie_fewshot}. It is also interesting to compare our proposed methods with recent image quality assessment approaches. For example, although aiming for a distinct objective of prioritising data to label, the proposed prioritisation metrics share technical similarities with previous work that quantifies task amenability of samples using direct feedback from a clinical task such as organ segmentation \cite{saeed_amenability, saeed_media, saeed_asmus, saeed_melba}. 
% In our previous work, we described a method to equip adaptability to the task amenability-predicting function \cite{saeed_media, saeed_asmus}. This work builds upon the same mechanism in order to make the AL system adaptable across different organs and images from different institutes.
Moreover, the proposed AL strategy is designed for medical images, as opposed to the language data used in previously proposed AL approaches that also utilised RL, e.g. Meng et al. \cite{meng_rl_active} and Woodward et al.~\cite{woodard_one_shot}, with algorithmic differences including problem definition, labelled example requirement, reward formulation and training methodology.

The contributions of this work are summarised as follows: 
1) We proposed a task-based AL metric with task-specific feedback from the targeted segmentation task; 
2) We proposed to learn the prioritisation metric using meta-RL with adaptability over different imaging institutes and organ segmentation tasks; 
3) We evaluated our proposed framework using real patient CT images and including segmentation tasks for anatomical structures such as liver, pancreas, spleen, liver vessels, gallbladder, adrenal glands (left and right), major vessels (aorta, vena cava and portal vein) and stomach; subsequently, the trained system was evaluated, for AL, on holdout tasks for liver vessels and kidneys for data from new institutes.

\section{Methods}

Pool sampling and stream sampling represent two typical cases for sampling the unlabelled data \cite{settles_active}, in which the data that need annotation become available sequentially or simultaneously, respectively. The sampling method is usually determined by the application. For example, in many medical image segmentation tasks, there are often more unlabelled images than labelled data. Considering the pool sampling in this study, when an unlabelled dataset is available, the so-called \textit{batch-mode} sampling \cite{settles_active} provides additional flexibility, with which the unlabelled images can be selected from and labelled in batches, hereinafter referred to as \textit{AL batches}, to allow efficient and practical parallel processing and annotating, with the AL batch size being an additional hyperparameter. 

We outline the task-based prioritisation metric formulation, in Sec. \ref{sec:bi_opt}, where the segmentation performance is quantified for a pool of images. We then outline an algorithm to equip adaptability to the metric, in Sec. \ref{sec:cont_train}, such that it can be adapted to new datasets. Finally, we describe the usage of the learnt metric for the batch-mode AL, in Sec. \ref{sec:act_learn}, during which the prioritisation function is used to select samples to label and train a task predictor for the new dataset. A high-level summary of all stages is also provided in Sec. \ref{sec:overview}.

\subsection{Preliminary: bi-level optimisation for task-based prioritisation}\label{sec:bi_opt}
\label{sec:method.preliminary}

In order to capture the task-specific performance for learning prioritisation, we consider two functions: 1) a \textit{task predictor} which performs the segmentation task; and 2) a \textit{controller} which predicts the task-specific prioritisation.

Defining the image and label domains for the task as $\mathcal{X}$ and $\mathcal{Y}$, respectively, we can denote the image and joint image-label distributions as $\mathcal{P}_{\mathcal{X}}$ and $\mathcal{P}_{\mathcal{X}\mathcal{Y}}$ with probability density functions $p(x)$ and $p(x,y)$, respectively. Here $x\in\mathcal{X}$ and $y\in\mathcal{Y}$ are the sampled image and corresponding label. We can then define our two functions as follows.

%\subsubsection{The task predictor function}

The task predictor is assumed to be a parametric function:
$f(\cdot; w): \mathcal{X} \rightarrow \mathcal{Y}$,
which outputs a prediction for the task $y\in\mathcal{Y}$ given an input image $x\in\mathcal{X}$, with parameters $w$. The controller is also assumed to be a parametric function:
$h(\cdot; \theta): \mathcal{X} \rightarrow [0,1]$,
which outputs a task-specific prioritisation score given an image sample $x$, with parameters $\theta$. To optimise the task predictor, a loss function $L_f: \mathcal{Y} \times \mathcal{Y} \rightarrow \mathbb{R}_{\geq0}$, which measures task performance, weighted by controller outputs $h(x;\theta)$, may be minimised:

\begin{align}\label{eq:predictor_opt}
\min_w \mathbb{E}_{(x,y)\sim\mathcal{P}_{XY}}[L_f(f(x;w), y)h(x;\theta)]
\end{align}
where, by weighting the loss using the controller outputs, the controller is incentivised to assign low scores to samples with high loss values.

The controller may be optimised by minimising a metric function $L_h: \mathcal{Y} \times \mathcal{Y} \rightarrow \mathbb{R}_{\geq0}$ on the validation set, weighted by the controller outputs for the validation set $h(x;\theta)$:

\begin{align}\label{eq:controller_opt}
\min_\theta \mathbb{E}_{(x,y)\sim\mathcal{P}_{XY}}[L_h(f(x;w), y)h(x;\theta)],\\
\text{s.t.}\quad \mathbb{E}_{x\sim\mathcal{P}_{X}}[h(x;\theta)] \geq C > 0.
\end{align}
where, higher metric function values tend the controller towards lower output values due to the weighted sum being minimised. The constraint, $C$, prevents $h\equiv0$ as a trivial solution.

% \subsubsection{Optimising both functions for prioritisation}

%\subsection{Bi-level optimisation for learning}

We can thus pose the following bi-level optimisation:

\begin{align}\label{eq:min_prob}
&& \min_\theta \mathbb{E}_{(x,y)\sim\mathcal{P}_{XY}}[L_h(f(x;w^*), y)h(x;\theta)],\\
\text{s.t.}&& w^*=\arg\min_w \mathbb{E}_{(x,y)\sim\mathcal{P}_{XY}}[L_f(f(x;w), y)h(x;\theta)],\\
&&\mathbb{E}_{x\sim\mathcal{P}_{X}}[h(x;\theta)] \geq C > 0.
\end{align}

Replacing the above functions weighted by the prioritisation scores with functions that sample only the selected images, retaining the equal expected function values, the optimisation problem becomes: 

\begin{align}\label{eq:min_prob_sampling}
&& \min_\theta \mathbb{E}_{(x,y)\sim\mathcal{P}_{XY}^h}[L_h(f(x;w^*), y)],\\
\text{s.t.}&& w^*=\arg\min_w \mathbb{E}_{(x,y)\sim\mathcal{P}_{XY}^h}[L_f(f(x;w), y)],\\
&&\mathbb{E}_{x\sim\mathcal{P}_{X}^h}[1] \geq C > 0.
\end{align}
where $x$ and $(x,y)$ to be sampled from the controller-selected or -sampled distributions $\mathcal{P}_{X}^h$ and $\mathcal{P}_{XY}^h$, with probability density functions $p^h(x) \propto p(x)h(x;\theta)$ and $p^h(x,y) \propto p(x,y)h(x;\theta)$, respectively. 

RL algorithms, previously proposed for task-specific image quality assessment \cite{saeed_amenability} or data valuation \cite{google_dvrl}, are adapted to optimise the task-based prioritisation metric. Building on this formulation, this work proposes an extension as well as a generalisation of such RL algorithms, a meta-RL approach for multiple datasets, as described in the remainder sections.

Once the controller is trained, it serves as a prioritisation function during AL. It is useful to clarify that, although the controller is denoted for individual samples for the simplicity in notation, its implementation includes a recurrent neural network (RNN), such that the episodic controller training (Algo.~\ref{algo:meta_iqa}) and the batch-mode sampling in AL stage (Algo. \ref{algo:act_learn}) enable learning and inferring any potentially useful sequential information, respectively. Such RNN-embedded RL agent has also been adopted in other meta-RL approaches \cite{duan_metarl, wang_metarl, botvinick_metarl, robles_metarl_nas}.

% \begin{figure}[!ht]
%     \centering
%     \includegraphics[width=0.49\textwidth, height=3.4cm]{meta_iqa_4.png}
%     \caption{Learning an adaptable prioritisation function. Learning happens across multiple environments each with a different segmentation task and data contained within the environment. Tasks are randomly sampled on each iteration for controller-environment interactions. Once an environment is sampled, a minibatch $\mathcal{B}_t$ is sampled from the data within the environment, which is passed to the controller in order to generate prioritisation scores/ selection probabilities for the samples. These are used to select samples (which is the controller action) to form $\mathcal{B}_{t, \text{selected}}$. $\mathcal{B}_{t, \text{selected}}$ is then used to update the task predictor. The reward used to train the controller comes from performance measured over the validation set $\tilde{R}_t=-\frac{1}{M}\sum_{j=1}^Ml_{j,t}h_j$. Another environment is then randomly sampled and the whole process is repeated. Eventually, the controller learns to prioritise samples based on the task-specific feedback from the reward signal.}
%     \label{fig:meta_iqa}
% \end{figure}

\subsection{Controller training stage: learning the adaptable prioritisation metric}\label{sec:cont_train}

In this work, the adaptability over different segmentation problems includes multi-organ adaptability as well as adaptability over different institutes, using the meta-RL-based training scheme described as follows.

\subsubsection{Markov decision process environment}

The above-outlined bi-level minimisation problem is modelled as a finite-horizon Markov decision process (MDP), where the controller interacts with an environment containing the task predictor and a specific set of data for training. The MDP-contained data are drawn from the distribution $\mathcal{P}_{XY}$, a joint image-label distribution, defined as $\mathcal{P}_{XY}= \mathcal{P}_X \mathcal{P}_{Y|X}$, with the task predictor as $f(\cdot;w)$. A train set $\mathcal{D}_{\text{train}} = \{ (x_i, y_i) \}_{i=1}^N$ is sampled from the distribution $\mathcal{P}_{XY}$, where $N$ is the train set size. The observed state of the environment $s_t = (f(\cdot; w_t), \mathcal{B}_t)$, at time-step $t$, is composed of the task predictor $f(\cdot; w)$ and a mini-batch of $b$ samples $\mathcal{B}_t = \{ (x_i, y_i) \}_{i=1}^b$ from the train set.

\subsubsection{Learning the metric using reinforcement learning}\label{sec:rl_iqa}

Reinforcement learning learns the prioritisation metric, by optimising the weights $\theta$ for the controller function $h(x;\theta)$. We define $\mathcal{S}$ as the state space and $\mathcal{A}$ as the continuous action space. $p: \mathcal{S} \times \mathcal{S} \times \mathcal{A} \rightarrow [0, 1]$ denotes the state transition distribution conditioned on state-actions, e.g. $p(s_{t+1} | s_t, a_t)$ represents the probability of the next state $s_{t+1} \in \mathcal{S}$ given the current state $s_t \in \mathcal{S}$ and action $a_t \in \mathcal{A}$. $r: \mathcal{S} \times \mathcal{A} \rightarrow \mathbb{R}$ is the reward function such that $R_t = r(s_t, a_t)$ denotes the reward given current state $s_t$ and action $a_t$. 

The policy, $\pi(a_t|s_t): \mathcal{S} \times \mathcal{A} \in [0, 1]$, represents the probability of performing the action $a_t$ given the state $s_t$. This allows for the MDP interaction to be summarised as a 5-tuple $(\mathcal{S}, \mathcal{A}, p, r, \pi)$. A number of interactions between the controller and the environments leads to a trajectory over multiple time-steps $(s_1, a_1, R_1, s_2, a_2, R_2, \ldots, s_T, a_T, R_T)$. The goal in this reinforcement learning is to optimise the controller parameters which maximise a cumulative reward over a trajectory. 

In our work, we use a cumulative reward, starting from time-step $t$, of the form: $Q^{\pi}(s_t, a_t) = \sum_{k=0}^{T} \gamma^k R_{t+k}$, where the discount factor $\gamma\in[0, 1]$ is used to discount future rewards. Here the policy is parameterised by $\theta$ and denoted as $\pi_{\theta}$. The central optimisation problem then is to find the optimal policy parameters $\theta^* = \text{argmax}_{\theta} \mathbb{E}_{\pi_{\theta}}\left[ Q^{\pi}(s_t, a_t) \right]$

Following Sec.\ref{sec:method.preliminary}, the controller outputs sampling probabilities $\{h(x_{i,t}, \theta)\}_{i=1}^b$ based on the input image batch. The action $a_t = \{a_{i,t}\}_{i=1}^b \in \{0, 1\}^b$ leads to a sample selection decision for task predictor training, selected if $a_{i,t}=1$. With $a_{i,t}\sim\text{Bernoulli}(h(x_{i,t}; \theta))$, the policy $\pi_{\theta}(a_t|s_t)$ is defined as:
    
\begin{align}
\log \pi_{\theta}(a_t|s_t) = \sum_{i=1}^b h(x_{i,t}; \theta) a_{i,t} + (1-h(x_{i,t}; \theta)(1-a_{i,t}))
\end{align}
    
The reward $R_t$ is based on the validation set $\mathcal{D}_{\text{val}} = \{(x_j, y_j)\}_{j=1}^M$ from the same distribution as the train set $\mathcal{P}_{XY}$, where $M$ is the validation set size. The performance for the validation set is denoted as $\{l_{j,t}\}_{j=1}^M = \{L_h(f(x_j;w_t), y_j)\}$ and is used to compute the un-clipped reward, which is weighted by the prioritisation scores $h_j$, $\tilde{R}_t=-\frac{1}{M}\sum_{j=1}^Ml_{j,t}h_j$. It is then clipped using a moving average, $\bar{R}_t ={\alpha}_R\bar{R}_{t-1}+(1-{\alpha}_R)\tilde{R}_t$, where ${\alpha}_R$ is a hyperparameter empirically set to 0.9. The final reward is then computed as $R_t = \tilde{R} - \bar{R}$.

This reward definition over a validation set, consisting of multiple samples, encourages a prioritisation metric to be learnt which promotes generalisability to new samples; this, in effect, captures a form of representativeness. At the same time, the task-specific nature of the reward signal allows for informativeness to be captured in the computed reward.

\subsubsection{A distribution of MDP environments}

We denote a distribution of MDP environments as $\mathcal{P}_M$. An MDP environment sampled from this distribution is thus denoted as $M_k \sim \mathcal{P}_M$. During training the controller, these MDP environments are collectively considered as the ``meta-train'' environments. An MDP environment is denoted using $M_k$ such that the the joint image-label distribution and task predictor within the environment can be denoted as $\mathcal{P}_{XY, k}$ and $f_{k}(\cdot;w_{k})$, respectively. 

In this work, the task predictor is a single neural network-based function approximator, shared across the MDP environments, denoted as $f(\cdot;w)$ by omitting $k$. The task predictor network are ``synced'' each time an environment is sampled, by updating the predictor parameters using a gradient-based meta-learning update step, detailed in the following section.

\subsubsection{Meta-RL to learn the adaptable prioritisation function}

Meta-RL aims to maximise the expected return over a distribution of environments, such that the trained controller may effectively adapt to new MDP environments sampled from the distribution \cite{wang_metarl, duan_metarl, botvinick_metarl}. 
The proposed episodic meta-RL training differs from RL in four aspects: 1) The controller is shared across multiple MDP environments sampled from $\mathcal{P}_M$; 2) Interaction of the controller with each MDP $M_k \sim \mathcal{P}_M$ takes place over multiple episodes and is referred to as a `trial'; 3) The controller embeds a RNN with the internal memory shared across episodes within the same trial. The RNN memory state is reset each time a new MDP is sampled. This enables adaptability with fixed weights, since the controller becomes a function of the history leading up to a sample of sequential input data; 4) The action $a_t$, raw reward $r_t$, and termination flag $d_t$ at the previous time step are passed to the controller as input, in addition to the observed current state $s_{t+1}$; the input denoted as $\tau_t$ encompasses these additional inputs. For per-sample controller operation, $r_t=R_t$ at the episode end and zero otherwise, i.e. a sparse reward \cite{duan_metarl, wang_metarl}.

In addition to an adaptable controller, the task predictor is shared between different environments and is updated using the Reptile update scheme \cite{nichol_reptile}, in order to equip adaptability to the task predictor as well. Whilst alternative gradient-based meta-learning algorithm may also be applicable, the Reptile-based task predictor update is efficient, formed of two steps: 1) Perform gradient descent for the task predictor $f(\cdot; w_t)$, starting with weights $w_t$ and ending in weights $w_{t, \text{new}}$; 2) Update the task predictor weights $w_t \leftarrow w_t + \epsilon(w_{t, \text{new}}-w_t)$. $\epsilon$ is set as 1.0 initially and linearly annealed to 0.0 as trials progress \cite{nichol_reptile}. We use adaptive moment estimation \cite{kingma_adam} as the gradient descent algorithm.

\begin{algorithm}[!ht]
\SetAlgoLined
\KwData{Multiple MDPs $M_k\sim\mathcal{P}_M$.}
\KwResult{Controller $h(\cdot;\theta)$.}
\BlankLine
\While{not converged}{
Sample an MDP $M_k\sim\mathcal{P}_M$\;
Reset the internal state of controller $h$\;
    \For{Each episode in all episodes}{
        \For{$t\leftarrow 1$ \KwTo $T$}{
            Sample a training mini-batch $\mathcal{B}_{t}=\{(x_{i,t},y_{i,t})\}_{i=1}^{b}$\;
            Compute selection probabilities $\{h_{i,t}\}_{i=1}^b=\{h(\tau_{i,t};\theta_t)\}_{i=1}^b$\;
            Sample actions $a_{t}=\{a_{i,t}\}_{i=1}^b$ w.r.t. $a_{i,t}\sim\text{Bernoulli}(h_{i,t})$\;
            Select samples $\mathcal{B}_{t,\text{selected}}$ from $\mathcal{B}_{t}$\;
            Update predictor $f(\cdot;w_t)$\ with $\mathcal{B}_{t,\text{selected}}$\;
            Compute reward $R_{t}$\;
        }
        Collect one episode $\{\mathcal{B}_t,a_t,R_t\}_{t=1}^T$\;

        Update controller $h(\cdot;\theta)$ using the RL algorithm described in Sec. \ref{sec:cont_train};
    }
}
\caption{Learning adaptable task-based prioritisation using multiple environments}
\label{algo:meta_iqa}
\end{algorithm}

Adaptation to a new MDP sampled from the distribution of MDPs, $M_a\sim\mathcal{P}_M$, is initiated by resetting the RNN internal state once at the start of the adaptation. The controller network weights are fixed, such that the adaptability comes from the RNN internal state updates rather than updates of the weights. This proposed scheme adapts the prioritisation metric to new datasets and, perhaps more importantly, enables a ``pre-trainable'' weight-fixed controller for the subsequent AL stage (Sec.~\ref{sec:act_learn}).

\begin{figure*}[!ht]
  \centering
  \subfloat[Learning an adaptable prioritisation function. Learning happens across multiple environments each with a different segmentation task and data contained within the environment. Tasks are randomly sampled on each iteration for controller-environment interactions. Once an environment is sampled, a minibatch $\mathcal{B}_t$ is sampled from the data within the environment, which is passed to the controller in order to generate prioritisation scores/ selection probabilities for the samples. These are used to select samples (which is the controller action) to form $\mathcal{B}_{t, \text{selected}}$. $\mathcal{B}_{t, \text{selected}}$ is then used to update the task predictor. The reward used to train the controller comes from performance measured over the validation set $\tilde{R}_t=-\frac{1}{M}\sum_{j=1}^Ml_{j,t}h_j$. Another environment is then randomly sampled and the whole process is repeated. Eventually, the controller learns to prioritise samples based on the task-specific feedback from the reward signal.]{\includegraphics[width=0.68\textwidth, height=3.4cm]{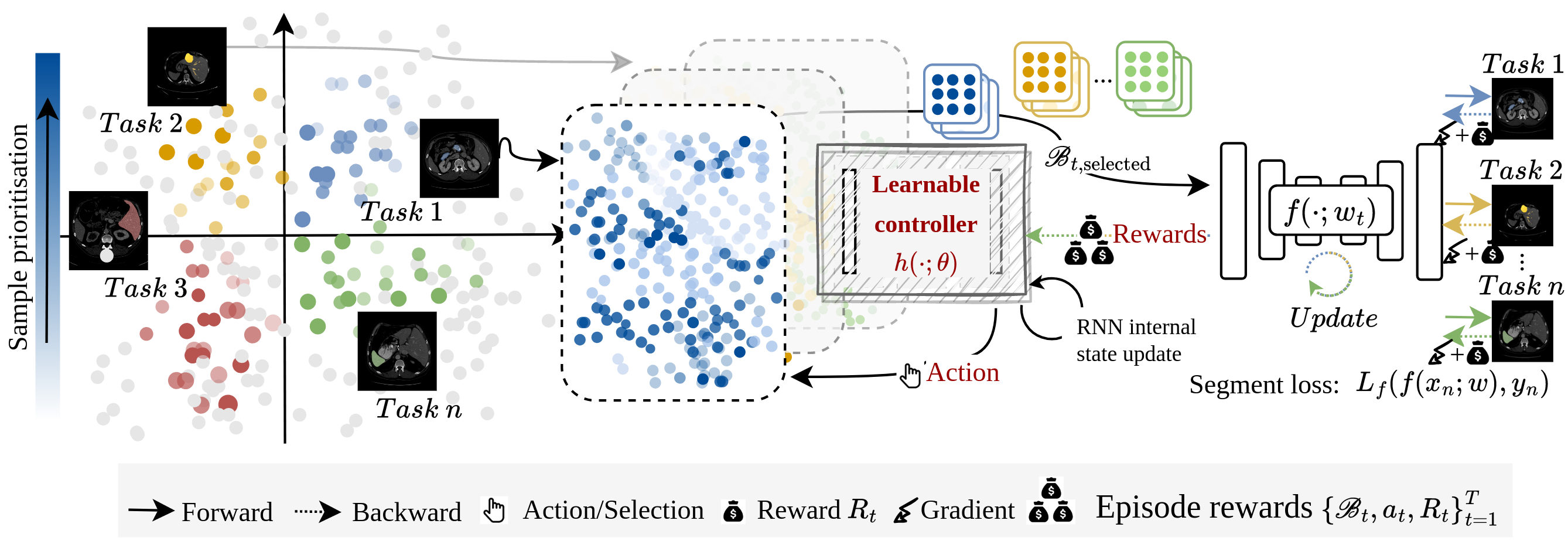}
    \label{fig:meta_iqa}}
  \hfill
  \subfloat[AL. Pre-trained controller (fixed weights), acting as a prioritisation function, outputs prioritisation scores. Highest priority samples are annotated by a human observer and form the support-train and support-validation sets, used to update the task predictor and internal state of the RNN-based controller for adaptation (using reward on the support-validation set as an explicit input), respectively.]{\includegraphics[width=0.26\textwidth, height=3.4cm]{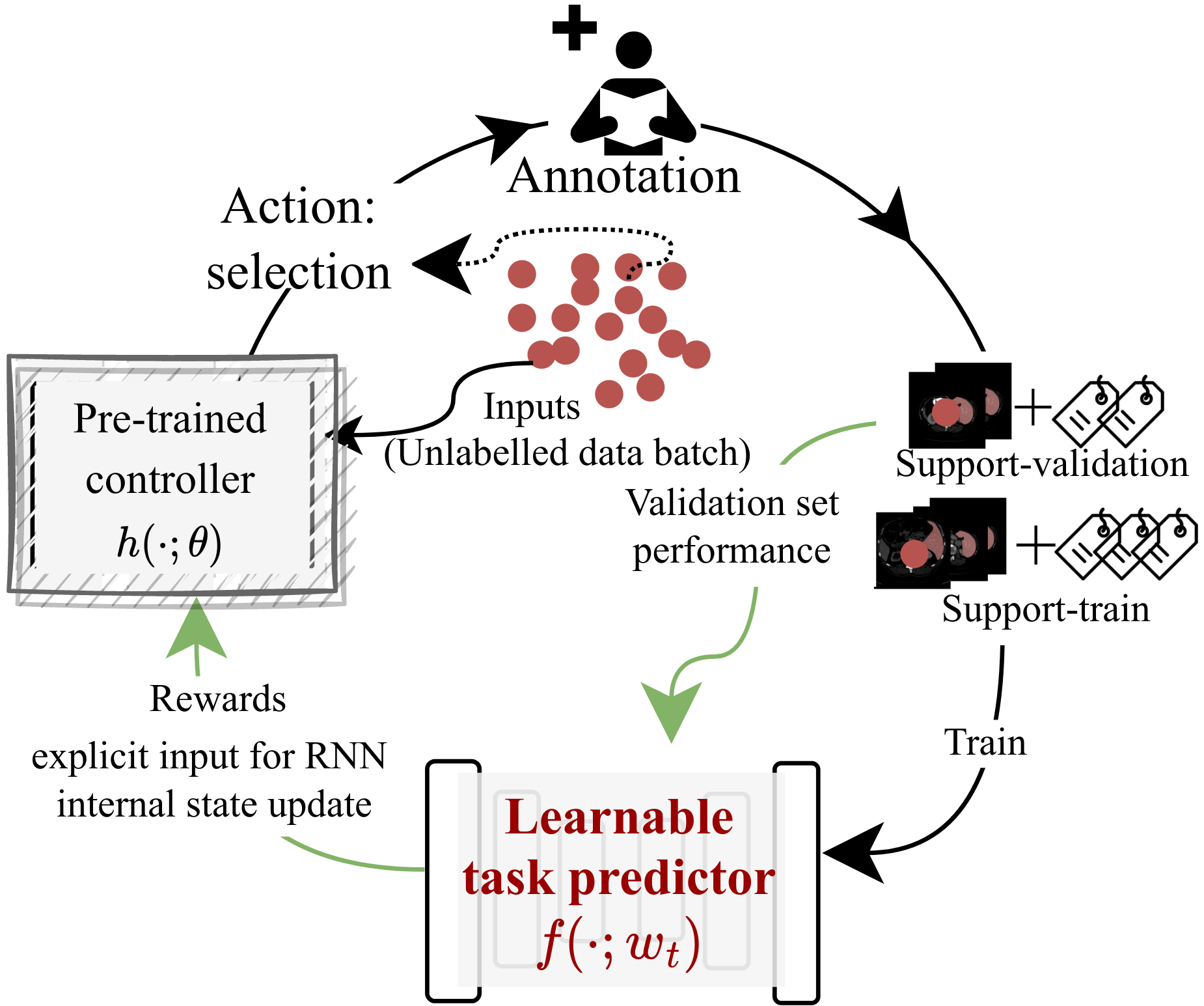}
    \label{fig:active_learn}}
\caption{Summary of two stages in the proposed framework.}
\end{figure*}

During the adaptation, the controller is adapted using the validation set for reward computation. The reward on the previous time-step is passed as an input explicitly to the controller to aid the above discussed adaptability. The detailed steps for leaning the adaptable task-based prioritisation are summarised in Algo. \ref{algo:meta_iqa} and illustrated in Fig.~\ref{fig:meta_iqa}.

\subsection{AL stage: using the pre-trained prioritisation function}\label{sec:act_learn}

The AL stage involves the use of a \textit{meta-test environment}, $M_a\sim\mathcal{P}_M$, which contains images from the joint image-label domain $\mathcal{P}_{XY}$, and is illustrated in Fig.~\ref{fig:active_learn}, with the algorithm summarised in Algo. \ref{algo:act_learn}. The entire dataset $\mathcal{D}$ is unlabelled in the initial meta-test environment and becomes labelled as AL iterates. The meta-train environments, however, do have labels. For brevity, it is not indexed by time steps, where more detailed data subsets are.

To initialise AL, we sample $\beta^0$ images randomly from the dataset $\mathcal{D} = \{x_i\}_{i=1}^{N+M}$, which consists of a pool of unlabelled samples, where sizes $N$ and $M$ are the intended sizes for the support-train\footnote{The ``support-'' prefix indicates the available datasets during AL stage and are processed by the weight-fixed controller. Readers familiar with meta-learning terminology may notice the absence of the ``query'' set, however, in our formulation, query set may be thought of as the entire pool of data that are available for AL, before any data annotation takes place, while increasing annotated samples are forming the growing support sets.} and support-validation sets, respectively, if the entire data pool is exhausted. These sampled $\beta^0$ images, labelled by an expert observer, form the initial support set $\mathcal{B}_{\text{support}} = \{x_i, y_i\}_{i=1}^{\beta^0}$. This support set is split into support-train set $\mathcal{B}_{\text{support-train}} = \{x_i, y_i\}_{i=1}^{\beta^0\times\phi}$ and a support-validation set $\mathcal{B}_{\text{support-validation}} = \{x_i, y_i\}_{i=1}^{\beta^0\times(1-\phi)}$, using a ratio of $\phi$. The support-train and support-validation sets are used to update the task predictor $f(\cdot; w)$ until convergence and to compute the reward, $R_{c}$, to form $\tau_{c+1, \cdot}$ which is passed to the controller as an explicit input on the next iteration, where $c=0$, respectively, in the initialisation step.

% \begin{figure}[!ht]
%     \centering
%     \includegraphics[width=0.4\textwidth]{active_5.png}
%     \caption{AL setup. The pre-trained controller (with fixed weights) serves as the prioritisation function during AL. The controller outputs prioritisation scores for samples in the dataset. The highest priority samples are then annotated by a human observer and are used to form the support-train and support-validation sets. These are used to update the task predictor and the internal state of the RNN-based controller to adapt it (using a reward computed on the support-validation set being used as an explicit input to the controller), respectively. }
%     \label{fig:active_learn}
% \end{figure}

With the remainder $N+M-\beta^0$ samples in the pool, the first AL iteration ($c=1$ counts the iterations) computes prioritisation scores using the fixed-weight pre-trained controller, $\{h(\tau_{i, c}; \theta^{*})\}_{i=1}^{N+M-\beta^0}$ where $\theta^*$ denotes pre-trained parameters, with only the RNN internal state adaptable. Among these unlabelled images $\{x_i\}_{i=1}^{N+M-\beta^0}$, the $\beta$ samples that are scored the highest by the prioritisation function are selected and labelled to form the support set at the iteration, $\mathcal{B}_{c, \text{support}} = \{x_i, y_i\}_{i=1}^{\beta}$, which is then further split to form the current support-train set $\mathcal{B}_{c, \text{support-train}} = \{x_i, y_i\}_{i=1}^{(\beta^0+\beta)\times (\phi)}$ and support-validation set $\mathcal{B}_{c, \text{support-validation}} = \{x_i, y_i\}_{i=1}^{(\beta^0+\beta)\times (1-\phi)}$, and added into respective support sets i.e. to $\mathcal{B}_{\text{support-train}}$ and $\mathcal{B}_{\text{support-validation}}$. The task predictor $f(\cdot; w)$ and the reward $R_{c}$ are updated using the support-train and support-validation sets, respectively, as described in the initialisation step. 

Thus in each subsequent $c^{th}$ iteration, $N+M-\beta^0-(\beta\times (c-1))$ images are available for prioritising and $\beta^0 + (\beta \times c)$ are labelled. The process is then repeated until the pool of unlabelled $N+M$ images is exhausted or until AL convergence is reached\footnote{AL convergence is different from the model convergence within each AL iteration, and refers to the AL system converging at the highest performance value across multiple AL iterations. In the context of AL with respect to the predictor, an AL iteration is complete when the learning system has converged to a performance value, given the data available at that particular instance.}. The number of samples for initialising AL $\beta^0$ may be different from $\beta$, since initialisation may require more samples compared to a single AL iteration \cite{budd_active_survey}.

\begin{algorithm}[!ht]
\SetAlgoLined
\KwData{MDP environment over which AL is to be conducted $M_a\sim\mathcal{P}_M$.}
\KwResult{Task predictor $f(\cdot;w)$.}
\BlankLine
Reset the internal state of controller $h$\;
From the dataset $\mathcal{D} = \{x_i\}_{i=1}^{N+M}$, randomly sample $\beta^0$ images and label them to form $\mathcal{B}_{\text{support}} = \{x_i, y_i\}_{i=1}^{\beta^0}$\;
Split $\mathcal{B}_{\text{support}}$ into support-train and support-validation portions $\mathcal{B}_{\text{support-train}} = \{x_i, y_i\}_{i=1}^{\beta^0\times (\phi)}$ and $\mathcal{B}_{\text{support-validation}} = \{x_i, y_i\}_{i=1}^{\beta^0\times (1-\phi)}$\;
Update the task predictor $f(\cdot; w)$ using $\mathcal{B}_{\text{support-train}}$\;
Compute the reward $R_{c}$ used to form $\tau_{c+1, \cdot}$ where $c=0$, using $\mathcal{B}_{\text{support-validation}}$\;
    \For{each AL iteration $c$ in all AL iterations}{
        \While{not converged}{
            Compute prioritisation scores for the remaining samples in the dataset $\{h(\tau_{i, c} ; \theta^{*})\}_{i=1}^{N+M-\beta^0-(\beta\times (c-1))}$\;
            Select $\beta$ high priority samples from the remaining samples in $\mathcal{D}$ i.e.  from $\{x_i\}_{i=1}^{N+M-\beta^0-(\beta\times (c-1))}$, label them and use them to form $\mathcal{B}_{c, \text{support}} = \{x_i, y_i\}_{i=1}^{\beta}$\;
            Split $\mathcal{B}_{c, \text{support}}$ into support-train and support-validation portions $\mathcal{B}_{c, \text{support-train}} = \{x_i, y_i\}_{i=1}^{\beta\times (\phi)}$ and $\mathcal{B}_{c, \text{support-validation}} = \{x_i, y_i\}_{i=1}^{\beta\times (1-\phi)}$ and add them to $\mathcal{B}_{\text{support-train}}$ and $\mathcal{B}_{\text{support-validation}}$\;
            Update predictor $f(\cdot;w_c)$ using $\mathcal{B}_{\text{support-train}}$\;
            Compute reward $R_{c}$ using $\mathcal{B}_{\text{support-validation}}$ and use it to form $\tau_{c+1, \cdot}$\;
        }
    }
\caption{AL using an adaptable prioritisation function (fixed controller weights)}
\label{algo:act_learn}
\end{algorithm}

\subsection{Stage summary for the proposed method}\label{sec:overview}

\begin{figure}[!ht]
    \centering
    \includegraphics[width=0.48\textwidth]{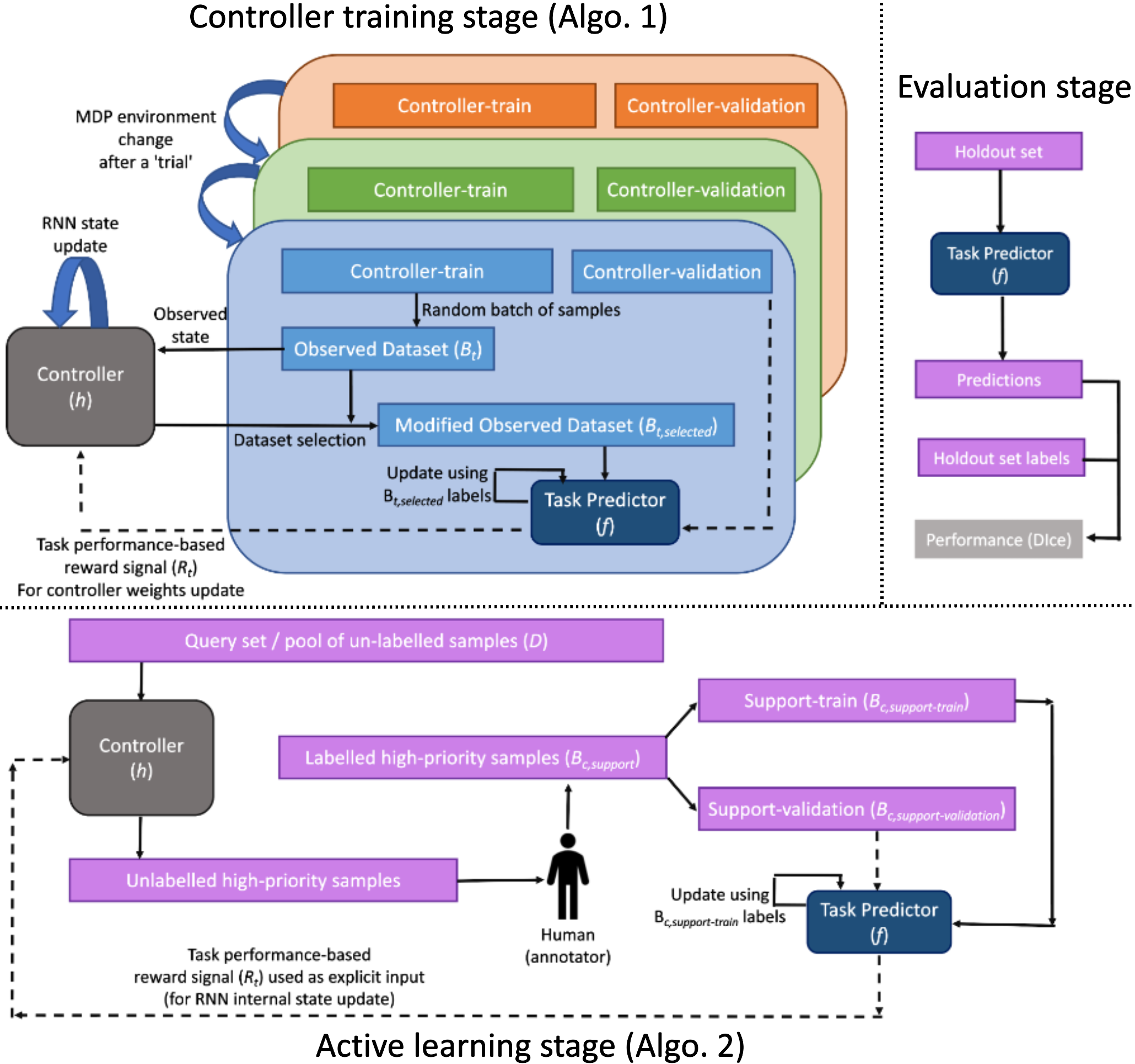}
    \caption{A high-level overview of the controller training, AL and evaluation stages (solid arrows indicate the train set pathway and dashed arrows indicate the validation set pathway).}
    \label{fig:overview}
\end{figure}

We now provide a high-level overview of all the stages involved including controller training (Algo. \ref{algo:meta_iqa}), AL (Algo. \ref{algo:act_learn}) and evaluation using the holdout set, with Fig. \ref{fig:overview}.

The controller training stage involves the use of multiple MDP environments, each with their own dataset. The dataset within each MDP includes the controller-train and controller-validation sets. One set of interactions with a single MDP environment or trial occurs when a random batch of samples is sampled from the controller-train set and passed to the controller to obtain prioritisation scores. Samples are then selected for training the task predictor using these scores (as outlined in Algo. \ref{algo:meta_iqa} using Reptile). Once trained, the task predictor is evaluated on the controller-weighted validation set (as described in Sec. \ref{sec:rl_iqa} and Algo. \ref{algo:meta_iqa}). The performance metric obtained from this evaluation is then used as a reward signal to update the controller. Then, a new MDP is randomly sampled and the controller updates continue until convergence. The controller is an RNN and the internal state is reset after each trial. It is also noteworthy that the controller takes additional inputs of the action, raw reward, and termination flag at the previous time step, in addition to the observed current state (batch of samples form the controller-train set) which makes it adaptable rather than any update of the weights.

For the AL stage, the controller weights are fixed after controller training and the internal state is reset before AL starts. Prior to AL, $\beta^0$ samples are randomly selected from a pool of unlabelled samples to initialise. These are labelled and split into support-train and support-validation portions. The support-train portion is used to updated the task predictor and the support validation portion is used to compute the reward which will be used as the controller input for the first AL iteration. The remaining samples in the pool of unlabelled samples (or the query set) are passed to the controller to obtain priority scores. $\beta$ highest priority samples are then selected and labelled. These labelled are then split into support-train and support-validation sets for the iteration and added to the respective support sets from the previous iteration, for their respective task predictor and reward updates. It should be noted that the reward in this case is used as controller input and not to update the controller weights as in the controller-training stage. The reward serves as a signal to adapt the RNN internal state which equips the system with adaptability without weight updates for the controller. The AL iterations continue until either exhausting the query set or convergence. 

To evaluate the AL performance, we use the task-predictor trained after AL. A holdout set, not used during the controller-training or AL stages, is passed to the trained task predictor. The performance measure is computed by comparing the task predictor predicted labels to the ground truth labels, where the mean binary Dice score is reported in our work. 

Code available at: github.com/s-sd/task-amenability/tree/v2

\vspace{-20pt}

\section{Experiments}\label{sec:exp}

In this section, we describe three experiments to assess the proposed controller, by evaluating the segmentation performance of kidney and liver vessels on abdominal CT images, as example anatomical structures for surgical planning interest. The clinical datasets used for controller training are described in Sec.~\ref{sec:exp.controller-train-data}; The three AL scenarios and the datasets are described in Sec.~\ref{sec:exp.AL-tasks}; The implementation is detailed in Sec.~\ref{sec:exp.implementation}; And the remainder Sec.~\ref{sec:exp.ablation} describes the alternative prioritisation metrics for ablation studies.

\subsection{Controller training datasets}
\label{sec:exp.controller-train-data}

The training data for the adaptable model comes from multiple institutes, summarised in Table \ref{tab:train_data}. The datasets in the table which have a cited reference as a source, are open-source datasets not requiring approval for usage; further ethical approval details for original acquisition may be found within the citations. Approval details for other datasets are provided in Sec. \ref{sec:exp.AL-tasks}. The datasets include organ segmentation tasks for the liver, pancreas, spleen, liver vessels, gallbladder, adrenal gland (left and right), major vessels (aorta, vena cava, portal and splenic veins), stomach and kidneys. Here, the `controller-' prefix indicates the use of the dataset during the controller training as outlined in Algo. \ref{algo:meta_iqa}. Two of the sets used for AL, i.e. those not used during controller training, are in-house datasets that come from UCL (University College London). It is noteworthy that the choice of annotation types were motivated by the intervention planning such as laparoscopic liver resection of interest in this study, as well as by the availability of the open data sets, which may further assist the re-producibility of the presented experimental results. As discussed in Sec.~\ref{sec:introduction}, combination of the training and support class types varies between applications and testing the ability to reduce the reliance on specific classes to be annotated for a large amount of training data is one of the focuses of the proposed cross-institute AL method.

% is it okay to add sources (open source links etc.) to an appendix and mention modality and organs here?

% maybe make a table here for all the data sources and the amount of data contained within each

\begin{table}[!ht]
\centering
\begin{tabular}{c c c c }
\hline
Source & Structure & Samples & Role \\
\hline
\cite{bilic_lits} \cite{antonelli_decathalon} & Liver & 131 & Training\\ %%
% \hline
\cite{antonelli_decathalon} & Pancreas & 281 & Training\\ %%
% \hline
\cite{antonelli_decathalon} & Spleen & 41 & Training\\ %%
% \hline
\cite{antonelli_decathalon} & Liver Vessels & 303 & Training\\ %%
% \hline
\cite{synapse_multi_atlas} & Gallbladder & 30 & Training\\ %%
% \hline
\cite{synapse_multi_atlas} & Adrenal Gland & 30 & Training\\ %%
% \hline
\cite{synapse_multi_atlas} & Major Vessels & 30 & Training\\ %%
% \hline
\cite{synapse_multi_atlas} & Stomach & 30 & Training\\ %%
% \hline
\cite{rister_ct_org} & Kidneys & 119 & AL\\ %%
% \hline
UCL & Kidneys & 35 & AL\\ %%
% \hline
UCL & Liver Vessels & 9 & AL\\ %%
\hline
\end{tabular}
\caption{Sources of training datasets used to form different environments. For the `Role', the `Training' means that two sets controller-train and controller-validation were created from the data; and `AL' means that the data was used to form the support-train and support-validation sets and another set called the holdout set, used to evaluate the AL performance.}
\label{tab:train_data}
\end{table}
\vspace{-10pt}

\subsection{Active learning tasks}
\label{sec:exp.AL-tasks}
The below described three \textit{meta-test} datasets were curated to represent the three AL scenarios found in clinical practice. These scenarios include the segmentation of ROIs that are from a different institute and/or completely unseen in controller-training. They are a mixture of openly accessible datasets from challenges and real-world clinical datasets from our local hospitals, as described in Table~\ref{tab:train_data}. In particular, the UCL data for vessel segmentation and kidney segmentation, Experiment (c) and (b), were acquired from patients undergoing laparoscopic liver resection surgery and interventional radiology for the kidneys, respectively also discussed in Sec.~\ref{sec:introduction}. In these surgical and interventional applications, the pre-operative data with the same acquisition protocols are inherently scarce, due to the prevalence of the surgical procedures at the local hospitals. Together with the other experiment with a much larger dataset, these tasks were designed to reflect the benefits, from the proposed prioritisation function, for scenarios with variable unlabelled data availability.
It should be noted that the $\phi$ parameter is simply the dataset split ratio of number of samples in support-train:support-validation. This is set empirically depending on the dataset size.

\paragraph{Cross-institute-and-organ kidney segmentation} Kidney segmentation data from Rister et al. \cite{rister_ct_org} is used as the first dataset for evaluation. This dataset contained 119 3D abdominal CT volumes with kidneys segmented manually for each volume. The data was split into 95, 12 and 12 samples in support-train, support-validation and holdout sets ($\phi=0.89$).

\paragraph{Cross-institute kidney segmentation} A second kidney segmentation task was used as another environment for evaluation, this dataset comprised of 35 3D abdominal CT scans from patients who presented with renal cancer and were undergoing renal cryoablation. Approval from the local clinical governance committee was obtained prior to data collection. Since this dataset is formed of CT scans for interventional use, different imaging protocols have been used compared to other sets used for training. The kidneys were segmented manually by a trained biomedical engineering researcher. The data was split into 21, 7 and 7 samples in the support-train, support-validation and holdout sets ($\phi=0.75$).

\paragraph{Cross-institute liver vessel segmentation} The third task used for evaluation is liver vessel segmentation. Nine 3D abdominal CT volumes (with liver vessels segmented using a commercial service \cite{visiblepatient}) were acquired in accordance with ethical standards of the institutional and/or national research committee and the 1964 Helsinki declaration (and amendments), under the study [REC=14/LO/1264][IRAS=158321]. Data was split into 3, 3 and 3 samples in support-train, support-validation and holdout sets ($\phi=0.5$).

% ct org for multi organ and multi centre - kidneys - 119; 95-12-12
% mark's data for multi centre - kidneys - 35; 21:7:7
% liver data - liver vessels - 9; 3:3:3

\subsection{Algorithm implementation}
\label{sec:exp.implementation}

A 3D U-Net \cite{3d_unet_seg} was used as the task predictor shared between environments. Dice loss (i.e. $1-\text{Dice}$) was used for training across all environments. The U-Net was formed of 4 down-sampling and 4 up-sampling layers where down-sampling modules consisted of two convolutional layers with batch normalisation and ReLU activation, and a max-pooling operation and de-convolution layers used in up-sampling modules. Each convolution layer doubles the number of channels where the number of channels for the first layer is 32. Encoding and decoding parts were connected using skip connections.

The algorithm used for training the controller was Proximal Policy Optimisation (PPO) \cite{schulman_ppo}. The actor and critic networks used in the PPO algorithm passes the image inputs via a 3-layered convolutional encoder which then feed into 3 fully connected layers, which embed an RNN. The reward used to train the controller was based on the Dice metric computed on the validation set. Using a single Nvidia Tesla V100 GPU, the controller training time was approximately 96 hours.

In this work AL experiments are set up as outlined in Sec. \ref{sec:act_learn}. First, $\beta^0$ samples are randomly chosen and labelled by an expert. Subsequently, $\beta$ additional samples are chosen on each new AL iteration, based on the prioritisation method used, and are labelled and used for further training together with samples from the previous AL iterations.

\subsection{Compared methods and ablation studies}
\label{sec:exp.ablation}

We use two prioritisation schemes in this work: 1) the proposed prioritisation scheme, which is the AL based adaptable prioritisation; and 2) a random prioritisation. We also use Monte-Carlo (MC) Dropout-based prioritisation to compare with our proposed prioritisation scheme and highlight this where comparisons are made. We compare the performance of the learning systems using t-tests and specify the AL iteration at which the comparison is being made, where appropriate. 

To evaluate the efficacy of the prioritisation metric, we perform an ablation study where the proposed controller-based adaptable prioritisation scheme is ablated from the proposed framework. This means that while the task predictor is the adaptable Reptile-based version, the prioritisation using the controller is replaced with random prioritisation or using an alternative prioritisation scheme. Results presented in Fig. \ref{fig:ablat}.

The reported performance is computed over a holdout set which is not used for training. The Dice score is reported for the learning systems over the holdout set. The settings for $\beta^0$ and $\beta$ are specified in Sec. \ref{sec:res}, where appropriate. Note that if $\beta\times (1-\phi) < 1.0$ then samples need not be added to the support-validation set on every AL iteration.

%\subsubsection{Network architectures}

\section{Results}\label{sec:res}

%The following subsections outline AL results for the three datasets used for AL described in Table \ref{tab:train_data}. Two prioritisation schemes are used for AL where random prioritisation serves as the baseline and the adaptive meta-RL-based prioritisation is our proposed prioritisation scheme.

\paragraph{Cross-institute and cross-organ adaptability for kidney segmentation}

% \begin{figure}[!ht]
%     \centering
%     \includegraphics[width=0.45\textwidth]{act_kid_1.png}
%     \caption{AL results for kidney segmentation using an open-source dataset with performance measured over a holdout set. ($n=24$, $m=4$)}
%     \label{fig:kidney_active_learn}
% \end{figure}

\begin{figure}[!ht]
  \subfloat[Cross-institute and cross-organ adaptability for kidney segmentation. \\($\beta^0=24$, $\beta=4$)]{\includegraphics[width=0.235\textwidth]{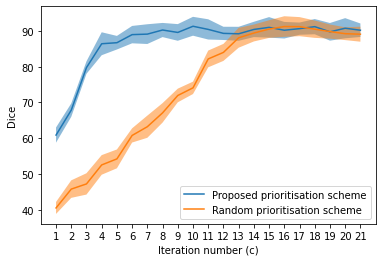}\label{fig:kidney_active_learn}}
  \hfill
  \subfloat[Cross-institute adaptability for kidney segmentation. \\($\beta^0=8$, $\beta=4$)]{\includegraphics[width=0.235\textwidth]{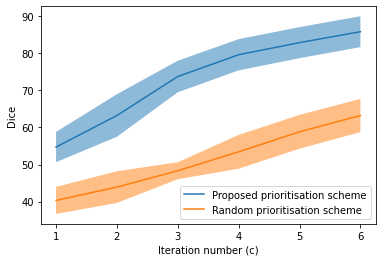}\label{fig:kidney_active_learn_2}}
  \vfill
    \subfloat[Cross-institute adaptability for \newline liver vessel segmentation. \\($\beta^0=2$, $\beta=1$)]{\includegraphics[width=0.235\textwidth]{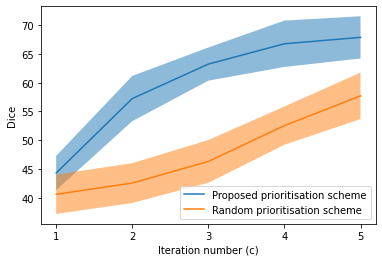}\label{fig:liver_active_learn}}
\hfill
  \subfloat[Ablation study for the kidney segmentation dataset. \\($\beta^0=24$, $\beta=4$)]{\includegraphics[width=0.235\textwidth]{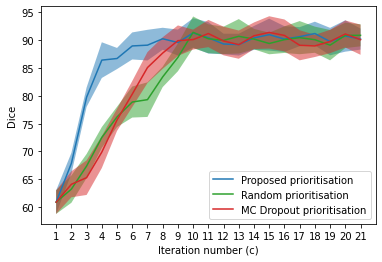}\label{fig:ablat}}
  \caption{AL performance over the holdout set.}
  \label{fig:all_res}
\end{figure}

As presented in Fig. \ref{fig:kidney_active_learn}, the proposed AL metric leads to faster convergence compared to random baseline with convergence reached near $c=6$ for the proposed prioritisation scheme compared to $c=12$ for the random prioritisation. It should be noted, however, that convergence is reached at a similar Dice score between these two ($p=0.09$, t-test at $\alpha$=0.05). While no difference was found between the two, both at $c=12$ ($p=0.11$), statistically significant difference was observed ($p<0.001$) between the random prioritisation with the proposed prioritisation scheme both at $c=6$.

\paragraph{Cross-institute adaptability for kidney segmentation}

% \begin{figure}[!ht]
%     \centering
%     \includegraphics[width=0.45\textwidth]{act_kid_2.png}
%     \caption{AL results for kidney segmentation using an in-house dataset with performance measured over a holdout set. ($n=8$, $m=4$)}
%     \label{fig:kidney_active_learn_2}
% \end{figure}

For the second kidney segmentation task, which demonstrates cross-institute-and-protocol adaptability, the plot of segmentation performance against AL iteration number is presented in Fig. \ref{fig:kidney_active_learn_2}. It appears as though convergence is not reached for either of the prioritisation schemes, however, at $c=6$ we observed an improvement in the Dice score for the proposed prioritisation scheme, with statistical significance ($p<0.001$). However, we observed higher Dice score for the proposed prioritisation metric for all values of $c$, with statistical significance ($p<0.001$ for all).

\paragraph{Cross-institute adaptability for liver vessel segmentation}

% \begin{figure}[!ht]
%     \centering
%     \includegraphics[width=0.45\textwidth]{act_liv_1.png}
%     \caption{AL results for liver vessel segmentation using an in-house dataset with performance measured over a holdout set. ($n=2$, $m=1$)}
%     \label{fig:liver_active_learn}
% \end{figure}

For the liver vessel experiments, we present the AL results in Fig. \ref{fig:liver_active_learn}. For this set of results, convergence may be considered inconclusive for either of the prioritisation schemes with the available data, however, we observe higher Dice score for the proposed prioritisation scheme, at $c=5$, compared with random prioritisation, with statistical significance ($p<0.001$). The proposed scheme results in higher performance compared to the random prioritisation baseline for all values of $c$, with statistical significance ($p<0.001$ for all).

\paragraph{Ablation study for the kidney segmentation dataset}

% \begin{figure}
%     \centering
%     \includegraphics[width=0.45\textwidth]{act_kid_1_ablat.png}
%     \caption{Ablation study results for kidney segmentation using an open-source dataset with performance measured over a holdout set; all model presented in this plot are initialised using the adaptable model and a different prioritisation scheme is used for each, subsequently. ($n=24$, $m=4$)}
%     \label{fig:ablat}
% \end{figure}

Comparing the ablated version with random prioritisation, a higher performance was seen from the proposed method from $c=2$ to $c=9$, with statistical significance ($p<0.001$ for all). Similarly, the proposed method also outperformed the Monte-Carlo Dropout-based prioritisation for $c=2$ to $c=8$, with statistical significance ($p<0.001$ for all). Interestingly, we found higher performance for the Monte-Carlo Dropout based scheme compared to random sampling for $c=7$ and $c=8$. The results are presented in Fig. \ref{fig:ablat}. All models presented in this plot are initialised using the adaptable model and a different prioritisation scheme is used for each, subsequently. Additionally, numerical results are presented in Tab. \ref{tab:res_tab_ablat}.

\begin{table}[]
\centering
\begin{tabular}{p{0.4cm} p{1.6cm} p{1.6cm} p{1.6cm} p{1.6cm}}
\hline
$c$ & Proposed & Random & Random \newline (adaptable task \newline predictor) & MC Dropout \newline (adaptable task \newline predictor) \\
\hline
1 & 60.89 $\pm$ 2.12 & 40.51 $\pm$ 1.67 & 60.89 $\pm$ 2.12 & 60.89 $\pm$ 2.12 \\ 

2 & 67.87 $\pm$ 1.87 & 45.78 $\pm$ 2.45 & 63.23 $\pm$ 2.46 & 64.10 $\pm$ 2.34 \\ 

3 & 79.76 $\pm$ 1.74 & 47.23 $\pm$ 2.98 & 67.38 $\pm$ 2.16 & 65.29 $\pm$ 3.11 \\ 

4 & 86.43 $\pm$ 3.21 & 52.54 $\pm$ 2.74 & 72.52 $\pm$ 1.87 & 69.89 $\pm$ 2.87 \\ 

5 & 86.72 $\pm$ 1.90 & 54.23 $\pm$ 2.63 & 76.22 $\pm$ 1.82 & 75.68 $\pm$ 1.98 \\ 

6 & 88.98 $\pm$ 2.43 & 60.79 $\pm$ 1.99 & 78.89 $\pm$ 2.79 & 80.32 $\pm$ 2.34 \\ 

7 & 89.12 $\pm$ 2.76 & 63.23 $\pm$ 3.10& 79.32 $\pm$ 3.11 & 85.09 $\pm$ 2.78 \\ 

8 & 90.27 $\pm$ 1.98 & 67.10 $\pm$ 2.72 & 83.41 $\pm$ 1.83 & 87.70 $\pm$ 2.84 \\ 

9 & 89.59 $\pm$ 2.33 & 71.96 $\pm$ 1.91 & 86.92 $\pm$ 2.49 & 89.93 $\pm$ 2.74 \\ 

10 & 91.32 $\pm$ 2.64 & 74.10 $\pm$ 1.73 & 91.43 $\pm$ 2.85 & 90.12 $\pm$ 1.79 \\ 

11 & 90.44 $\pm$ 2.83 & 82.18 $\pm$ 2.34 & 90.27 $\pm$ 2.60 & 91.21 $\pm$ 2.45 \\ 

12 & 89.33 $\pm$ 1.85 & 84.01 $\pm$ 2.38 & 89.98 $\pm$ 2.46 & 89.87 $\pm$ 2.58 \\ 

13 & 89.21 $\pm$ 1.96 & 88.10 $\pm$ 2.78 & 90.71 $\pm$ 3.10 & 89.21 $\pm$ 2.54 \\ 

14 & 90.43 $\pm$ 2.14 & 89.43 $\pm$ 2.36 & 90.21 $\pm$ 1.84 & 90.81 $\pm$ 2.39 \\ 

15 & 91.01 $\pm$ 2.87 & 90.53 $\pm$ 2.47 & 89.42 $\pm$ 1.94 & 91.41 $\pm$ 2.91 \\ 

16 & 90.21 $\pm$ 2.32 & 91.24 $\pm$ 2.88 & 90.11 $\pm$ 2.39 & 90.87 $\pm$ 2.87 \\ 

17 & 90.65 $\pm$ 1.71 & 91.21 $\pm$ 2.67 & 90.47 $\pm$ 2.98 & 89.10 $\pm$ 2.71 \\ 

18 & 91.21 $\pm$ 2.13 & 90.50 $\pm$ 2.43 & 90.10 $\pm$ 2.35 & 88.98 $\pm$ 1.99 \\ 

19 & 89.76 $\pm$ 2.47 & 89.83 $\pm$ 1.98 & 89.12 $\pm$ 2.73 & 89.78 $\pm$ 2.10 \\ 

20 & 90.77 $\pm$ 2.82 & 89.24 $\pm$ 1.79 & 90.88 $\pm$ 2.28 & 91.13 $\pm$ 2.43 \\ 

21 & 90.21 $\pm$ 1.89 & 89.10 $\pm$ 2.10 & 90.89 $\pm$ 1.98 & 90.10 $\pm$ 2.71 \\ 
\hline
\end{tabular}
\caption{AL performance (Dice) measured over the holdout set for different prioritisation methods.}
\label{tab:res_tab_ablat}
\end{table}
\vspace{-20pt}

\vspace{-10pt}

\paragraph{Exploring the impact of hyperparameters}

\begin{figure}[!ht]
  \centering
  \subfloat[$\beta^0$]{\includegraphics[width=0.23\textwidth]{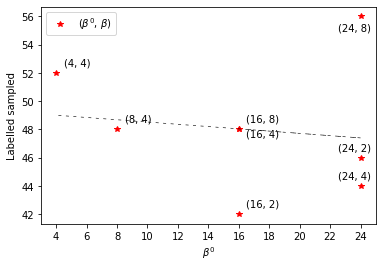}
    \label{fig:hyperparam_beta_0}}
  \hfill
  \subfloat[$\beta$]{\includegraphics[width=0.23\textwidth]{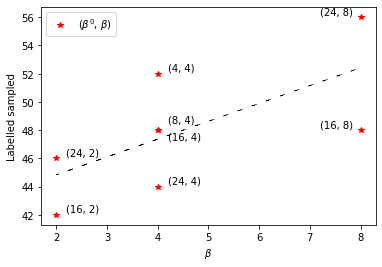}
    \label{fig:hyperparam_beta}}
\caption{Total number of labelled samples for convergence against hyperparameter, with fitted straight lines for reference.}
\label{fig:hyperparam}
\end{figure}

Eight hyperparameter combinations (different values of $\beta$ and $\beta^0$) were tested for the kidney segmentation task for the proposed active sampling. Consistent conclusions can be drawn with AL convergence being reached using nearly the same number of labelled samples as in our experiments outlined above. Detailed results are presented in Tab. \ref{tab:hyperparam}. Fig. \ref{fig:hyperparam} illustrates final performance and number of samples required to reach convergence, relatively insensitive to varying hyperparameters. 

\begin{table}[!ht]
\centering
\begin{tabular}{c c c c c}
\hline
$\beta^0$ & $\beta$ & Dice & $c$ at convergence & Labelled samples\\
\hline
24 & 4  & 88.98 $\pm$ 2.34 &  5 & 44 \\
16 & 4  & 90.13 $\pm$ 1.83 &  8 & 48 \\
8 & 4   & 91.02 $\pm$ 2.67 & 10 & 48 \\
4 & 4   & 89.95 $\pm$ 1.94 & 12 & 52 \\
24 & 2  & 88.87 $\pm$ 2.17 & 11 & 46 \\
24 & 8  & 90.68 $\pm$ 1.75 &  4 & 56 \\
16 & 2  & 89.40 $\pm$ 2.24 & 13 & 42 \\
16 & 8  & 91.11 $\pm$ 1.99 &  4 & 48 \\
\hline
\end{tabular}
\caption{hyperparameters tested for the kidney task.}
\label{tab:hyperparam}
\end{table}

\vspace{-20pt}

\paragraph{Analysis of the learnt selection strategies}

\begin{figure}[!ht]
  \centering
  \subfloat[Support sets by random vs by the proposed ]{\includegraphics[width=0.156\textwidth]{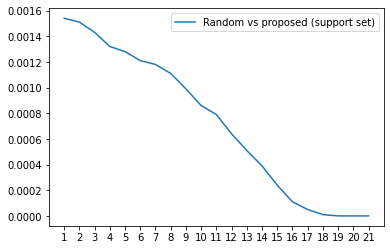}\label{fig:mmd_rand_prop}}
  \hfill
  \subfloat[Support set vs entire pool]{\includegraphics[width=0.156\textwidth]{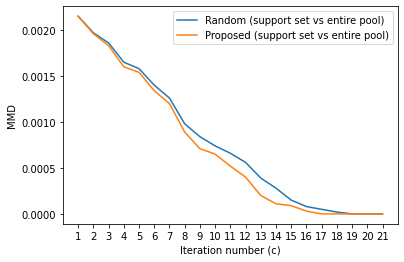}\label{fig:mmd_entire}}
    \hfill
  \subfloat[Support set vs holdout set (text for details)]{\includegraphics[width=0.156\textwidth]{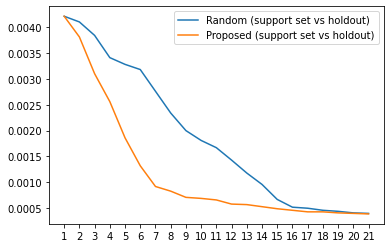}\label{fig:mmd_holdout}}
  \caption{MMD against iteration number.}
  \label{fig:mmds}
\end{figure}

To further analyse the selected samples at any given iteration, the Maximum Mean Discrepancy (MMD) was computed between $\mathcal{B}_{\text{support}}$ images selected by the two sampling strategies i.e. random and the proposed prioritisation, with the MMD value against the iteration number plotted in Fig. \ref{fig:mmd_rand_prop}. MMD is also computed between the support set at iteration $c$ and the entire pool of available data, plotted in Fig. \ref{fig:mmd_entire}. The random and proposed prioritisation follow a indistinguishable decreasing pattern, which may suggest the inability of MMD itself to differentiate or prioritise. 
%We hypothesise that since the support set is a subset of the entire set for both schemes, the MMD may not accurately represent the difference between the sampling schemes due to it being heavily influenced by the same samples in both sets. 
Finally, MMD computed between the support set at iteration $c$ and the holdout set is presented in Fig. \ref{fig:mmd_holdout}. In contrast, we observed faster decline in MMD for the proposed scheme in this case. The more observable difference may be due to the fact that the support is not a subset of the holdout and thus the divergence can be estimated more accurately using MMD. The difference itself may suggest a more representative selection learnt from the proposed method.

% \begin{figure}
%     \centering
%     \includegraphics[width=0.24\textwidth]{mmds.png}
%     \caption{MMD against iteration number computed between the support sets selected by the random and proposed prioritisation schemes.}
%     \label{fig:mmd}
% \end{figure}

\paragraph{Qualitative analysis of results}

\begin{figure}[!ht]
    \centering
    \includegraphics[width=0.48\textwidth]{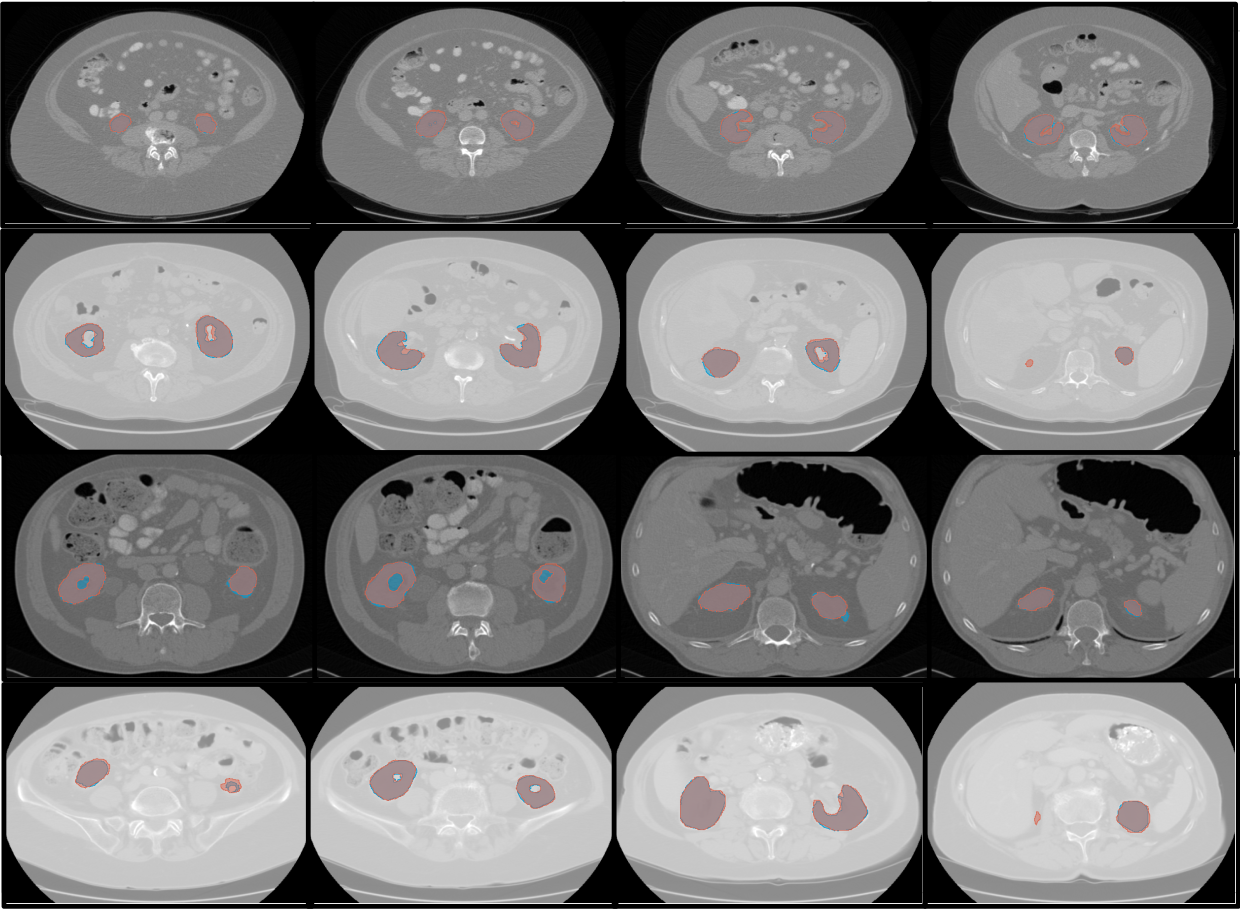}
    \caption{Samples of kidney segmentation with ground truth (blue) and prediction (red), rows indicating same patients.}
    \label{fig:kidney_samples}
\end{figure}

\begin{figure}[!ht]
  \centering
  \subfloat[Dice = 0.75]{\includegraphics[width=0.24\textwidth]{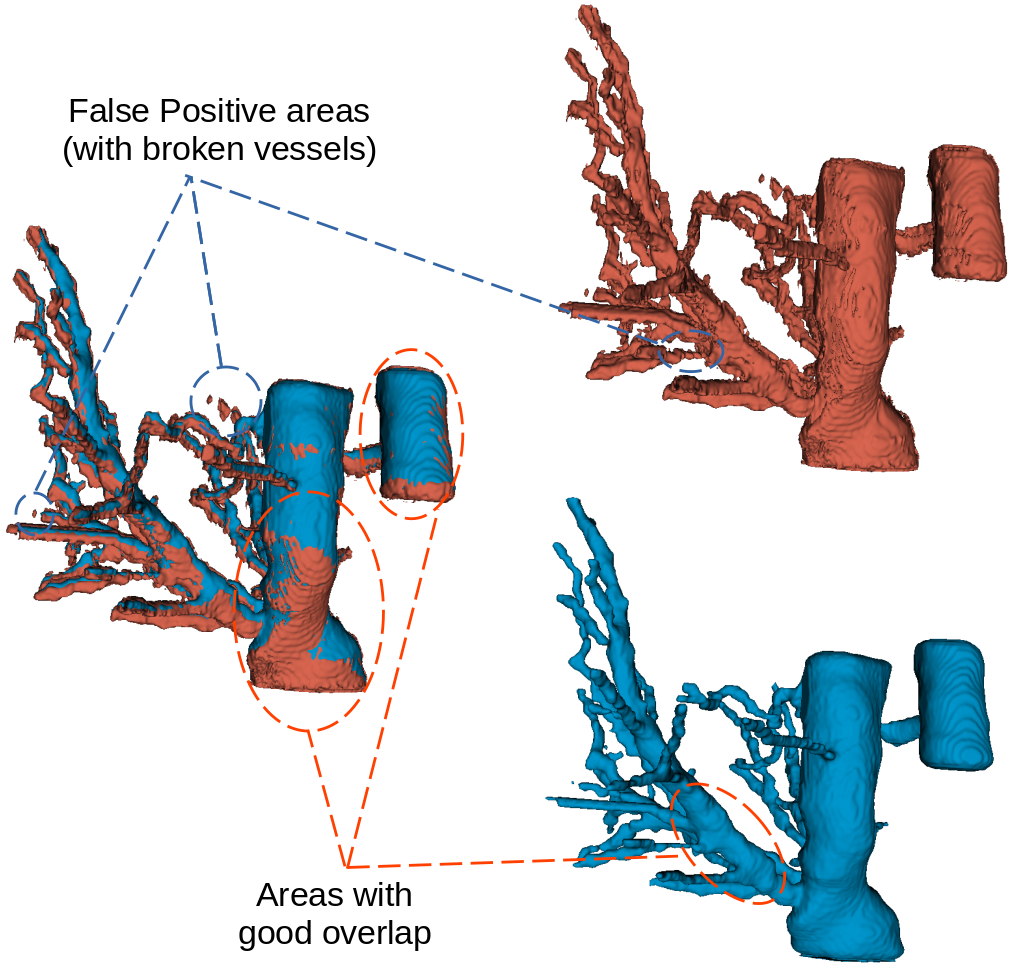}\label{fig:liv_vessel_model}}
  \hfill
  \subfloat[Dice = 0.64]{\includegraphics[width=0.24\textwidth]{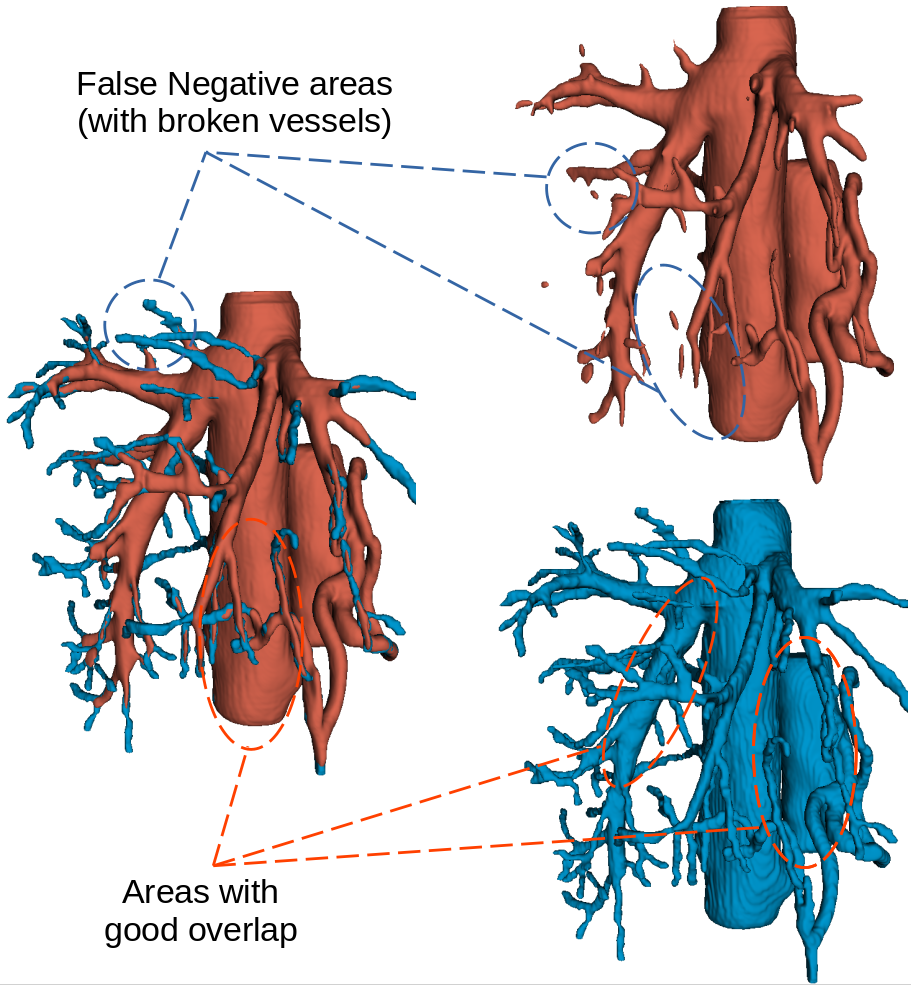}\label{fig:liv_vessel_model_2}}
  \caption{3D rendering of ground truth (blue) and overlaid prediction (red) for liver vessel task, from two holdout patients.}
  \label{fig:liv_vessel}
\end{figure}

\begin{figure}
    \centering
    \includegraphics[width=0.48\textwidth]{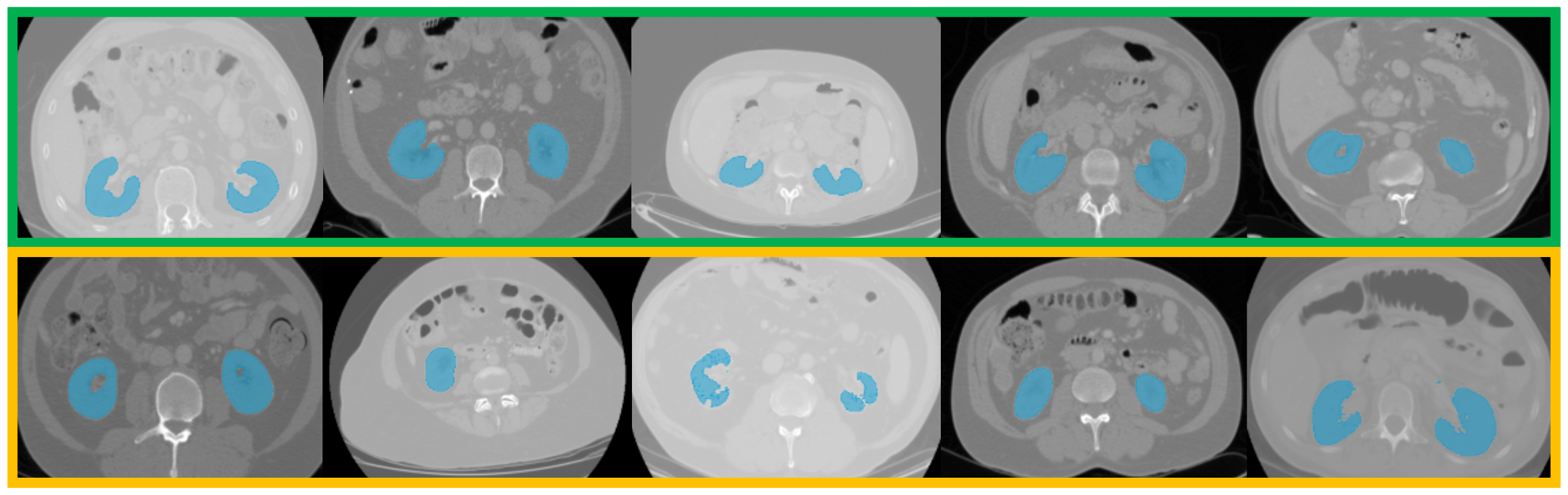}
    \caption{Example samples selected by the proposed (green) and random prioritisation (yellow) for the kidney task ($c=5$).}
    \label{fig:rand_prior_comp}
\end{figure}

Fig. \ref{fig:kidney_samples} shows examples of segmented kidneys from the holdout dataset.
Fig. \ref{fig:liv_vessel} presents 3D models of liver vessels used as ground truth and corresponding prediction form the trained AL model.
Fig. \ref{fig:rand_prior_comp} compares examples selected by the proposed and random prioritisation schemes. It can be seen that randomly selected samples may include ``low quality'' or less-representative examples, for example, a sample with a missing label for one kidney (second) and a sample with disconnections in labels (third), while these cases are much fewer in samples selected by proposed prioritisation.

% \begin{figure}[!ht]
%     \centering
%     \includegraphics[width=0.48\textwidth]{liv_vessel_model.png}
%     \caption{A 3D model/ rendering of the ground truth data and the corresponding model prediction for the liver vessel segmentation task used in the experiment. Green is the ground truth segmentation and red is the model-predicted segmentation. (Dice=0.75)}
%     \label{fig:liv_vessel_model}
% \end{figure}

% \begin{figure}[!ht]
%     \centering
%     \includegraphics[width=0.48\textwidth]{liv_vessel_model_2.png}
%     \caption{A 3D model/ rendering of the ground truth data and the corresponding model prediction for the liver vessel segmentation task used in the experiment. Green is the ground truth segmentation and red is the model-predicted segmentation. (Dice=0.64)}
%     \label{fig:liv_vessel_model_2}
% \end{figure}

\vspace{-30pt}

\vspace{-20pt}

\section{Discussion}

% faster convergence for proposed scheme
% ablation shows that even just the prioritisation scheme is effective and it is not only because of the adaptable task predictor
% can be used for multiple centres and organs in the abdomen
% performance improvement is not seen but faster convergence is seen with saving in terms of samples
% potential to use AL for new datasets.

%Results presented in Sec. \ref{sec:res} show that the AL framework can effectively reduce the number of samples required to reach convergence or to reach a higher performance level compared to a random prioritisation scheme. We demonstrate this using three datasests for two tasks including kidney segmentation and liver vessel segmentation. This shows that the combined adaptable task predictor model together with the adaptable prioritisation are more effective compared with random prioritisation.

Results from Sec. \ref{sec:res} show that the AL can effectively reduce the number of samples required to reach convergence, or a higher performance level, compared to random prioritisation. This means, to achieve the same performance, experts would need to label fewer samples compared to annotating randomly sampled images. We demonstrate this using three datasests for two tasks: kidney segmentation and liver vessel segmentation. For kidney segmentation, the adaptable prioritisation metric yields converging segmentation accuracy using only 40-60\% of labels otherwise required with other prioritisation metrics or random sampling. For datasets with limited size, for kidney and liver vessel segmentation, the adaptable prioritisation metric offers a performance improvement of 22.6\% and 10.2\% in Dice score, respectively, compared to random prioritisation. This directly corresponds to savings in clinician time. 

It is important to clarify that, based on the presented ablation results, the proposed adaptable prioritisation outperformed other schemes in the number of AL iterations to convergence. The performance at convergence, however, is comparable for all tested variants. Performance improvement is arguably not the goal of AL.
%We demonstrated the effectiveness of the model for the segmentation of a new organ, the kidney, which was not used during training and for data from a new institute, not used during training for tasks of kidney segmentation and liver vessel segmentation. This demonstrates the potential to use such AL systems for new datasets where abdominal organ segmentation on 3D CT scans needs to be performed. A saving in the number of labelled samples required to train an automated segmentation model that reaches the same performance as a model trained using all the data allows for savings in terms of the clinician's time and in terms of the effort required to train such an automated system.
However, the proposed adaptive AL has potential for a wider range of scenarios such as adaptable AL metrics across different tasks, across different observers and their labelling protocols, in addition to what are demonstrated in this paper, for novel structures and imaging institutes.

\section{Conclusion}

This work introduces an adaptable AL metric learnt using a meta-RL approach which uses direct task-specific feedback for labelling prioritisation. The proposed method leads to faster convergence compared to random prioritisation and the widely used Monte-Carlo Dropout-based method. We demonstrated the applicability of the proposed approach on three segmentation tasks using multi-organ multi-institute CT data.

\section*{Acknowledgements}

This work is supported by the EPSRC grant [EP/T029404/1]; Wellcome/EPSRC Centre for Interventional and Surgical Sciences [203145Z/16/Z]; the NIHR i4i grant [II-LA-1116-20005] and the EPSRC CDT in i4health [EP/S021930/1].

\bibliographystyle{IEEEtran}
\bibliography{bibliography}

\end{document}